\documentclass[11pt,a4paper]{article}
\usepackage[hyperref]{acl2018}
\usepackage{times}
\usepackage{latexsym}
\usepackage{caption}
\usepackage{url}
\usepackage{enumitem}
\usepackage{graphicx}
\usepackage{pgfplots}
\usepackage{amsmath}

\pgfplotsset{width=7cm,compat=1.8}
\usepackage{array}
\usepackage[compact]{titlesec}
\let\ACMmaketitle=\maketitle
\renewcommand{\maketitle}{\begingroup\let\footnote=\thanks \ACMmaketitle\endgroup}
\newcolumntype{M}[1]{>{\centering\arraybackslash}m{#1}}
\aclfinalcopy 

\title{TutorialBank:
A Manually-Collected Corpus for Prerequisite \\Chains,  Survey Extraction and Resource Recommendation}

\author{Alexander R. Fabbri \qquad Irene Li \qquad Prawat Trairatvorakul \qquad Yijiao He  \\   \textbf{Wei Tai Ting}  \quad \hspace{0.01\textwidth} \textbf{Robert Tung}  \quad \hspace{0.01\textwidth} \textbf{Caitlin Westerfield}   \quad \hspace{0.01\textwidth} \textbf{Dragomir R. Radev}  \AND
  {\normalfont Department of Computer Science, Yale University} \\
\scalebox{0.80}[0.9]{{ \{alexander.fabbri,irene.li,prawat.trairatvorakul,yijiao.he,}}\\\scalebox{0.80}[0.9]{{ robert.tung,weitai.ting,caitlin.westerfield,dragomir.radev\}@yale.edu}} \AND \vspace{-1.6cm}}

\date{}

\begin{document}
\maketitle
\begin{abstract}
The field of Natural Language Processing (NLP) is growing rapidly, with new research published daily along with an abundance of tutorials, codebases and other online resources. In order to learn this dynamic field or stay up-to-date on the latest research, students as well as educators and researchers must constantly sift through multiple sources to find valuable, relevant information. To address this situation, we introduce TutorialBank, a new, publicly available dataset which aims to facilitate NLP education and research. We have manually collected and categorized over 6,300 resources on NLP as well as the related fields of Artificial Intelligence (AI), Machine Learning (ML) and Information Retrieval (IR). Our dataset is notably the largest manually-picked corpus of resources intended for NLP education which does not include only academic papers. Additionally, we have created both a search engine \footnote{\url{http://aan.how}} and a command-line tool for the resources and have annotated the corpus to include lists of research topics, relevant resources for each topic, prerequisite relations among topics, relevant sub-parts of individual resources, among other annotations. We are releasing the dataset and present several avenues for further research.  
\end{abstract}

\section{Introduction}

NLP has seen rapid growth over recent years. A Google search of ``Natural Language Processing" returns over 100 million hits with papers, tutorials, blog posts, codebases and other related online resources. Additionally, advances in related fields such as Artificial Intelligence and Deep Learning are strongly influencing current NLP research. With these developments, an increasing number of tutorials and online references are being published daily. As a result, the task of students, educators and researchers of tracking the changing landscape in this field has become increasingly difficult.
\par

Recent work has studied the educational aspect of mining text for presenting scientific topics. One goal has been to develop concept maps of topics, graphs showing which topics are prerequisites for learning a given topic \cite{Gordon:16, Liu:16, Pan:17, Pan:17b, Liang:17}. Another goal has been to automatically create reading lists for a subject either by building upon concept graphs \cite{Gordon:17} or through an unstructured approach \cite{Jardine:14}.
\par
Additionally, other work has aimed to automatically summarize scientific topics, either by extractively summarizing academic papers \cite{Jha:13, Jha:15, Jaidka:16} or by producing Wikipedia articles on these topics from multiple sources \cite{Sauper:09, Liu:18}. Scientific articles constitute primary texts which describe an author's work on a particular subject, while Wikipedia articles can be viewed as tertiary sources which summarize both results from primary works as well as explanations from secondary sources. \newcite{Tang:04, Tang:09} and \newcite{Sheng:17} explore the pedagogical function among the types of sources. 
\par
To address the problem of the scientific education of NLP more directly, we focus on the annotation and utilization of secondary sources presented in a manner immediately useful to the NLP community. We introduce the TutorialBank corpus, a manually-collected dataset of links to over 6,300 high-quality resources on NLP and related fields. The corpus's magnitude, manual collection and focus on annotation for education in addition to research differentiates it from other corpora. Throughout this paper we use the general term ``resource" to describe any tutorial, research survey, blog post, codebase or other online source with a focus on educating on a particular subject. We have created a search engine for these resources and have annotated them according to a taxonomy to facilitate their sharing. Additionally, we have annotated for pedagogical role, prerequisite relations and relevance of resources to hand-selected topics and provide a command-line interface for our annotations. 

\par
Our main contribution is the manual collection of good quality resources related to NLP and the annotation and presentation of these resources in a manner conducive to NLP education. Additionally, we show initial work on topic modeling and resource recommendation. We present a variant of standard reading-list generation which recommends resources based on a title and abstract pair and demonstrate additional uses and research directions for the corpus.   

\section{Related Work}
\subsection{Pedagogical Value of Resources}
Online resources are found in formats which vary in their roles in education. \newcite{Sheng:17} identify seven types of pedagogical roles found in technical works: Tutorial, Survey, Software Manual, Resource, Reference Work, Empirical Results, and Other. They annotate a dataset of over 1,000 resources according to these types. Beyond these types, resources differ in their pedagogical value, which they define as ``the estimate of how useful a document is to an individual who seeks to learn about specific concepts described in the document". \newcite{Tang:04, Tang:09} discuss the pedagogical value of a single type, academic papers, in relation to a larger recommendation system. 
\subsection{Prerequisite Chains} 
Prerequisite chains refer to edges in a graph describing which topics are dependent on the knowledge of another topic. Prerequisite chains play an important role in curriculum planning and reading list generation. \newcite{Liu:16} propose ``Concept Graph Learning" in order to induce a graph from which they can predict prerequisite relations among university courses. Their framework consists of two graphs: (1) a higher-level graph which consists of university courses and (2) a lower-level graph which consists of induced concepts and pair-wise sequential preferences in learning or teaching the concept. 
\par
\newcite{Liang:17} experiment with prerequisite chains on education data but focus on the recovery of a concept graph rather than on predicting unseen course relations as in \newcite{Liu:16}. They introduce both a synthetic dataset as well as one scraped from 11 universities which includes course prerequisites as well as concept-prerequisite labels. Concept graphs are also used in \cite{Gordon:16} to address the problem of developing reading lists for students. The concept graph in this case is a labeled graph where nodes represent both documents and concepts (determined using Latent Dirichlet Allocation (LDA) \cite{Blei:03}), and edges represent dependencies. They propose methods based on cross entropy and information flow for determining edges in the graph. Finally, finding prerequisite relationships has also been used in other contexts such as Massive Open Online Courses (MOOCs) \cite{Pan:17,Pan:17b}.

\subsection{Reading List Generation}
\newcite{Jardine:14} generates recommended reading lists from a corpus of technical papers in an unstructured manner in which a topic model weighs the relevant topics and relevant papers are chosen through his ThemedPageRank approach. He also provides a set of expert-generated reading lists. Conversely, \newcite{Gordon:17} approach reading list generation from a structured perspective, first generating a concept graph from the corpus and then traversing the graph to select the most relevant document. 
\subsection{Survey Extraction}
Recent work on survey generation for scientific topics has focused on creating summaries from academic papers \cite{Jha:13, Jha:15, Jaidka:16}. \newcite{Jha:13} present a system that generates summaries given a topic keyword. From a base corpus of papers found by query matching, they expand the corpus via a citation network using a heuristic called Restricted Expansion.  This process is repeated for seven standard NLP topics. In a similar manner, \newcite{Jha:15} experiment with fifteen topics in computational linguistics and collect at least surveys written by experts on each topic, also making use of citation networks to expand their corpus. They introduce a content model as well as a discourse model and perform a qualitative comparisons of coherence with a standard summarization model.  
\par
The task of creating surveys for specified topics has also been viewed in the multi-document summarization setting of generating Wikipedia articles \cite{Sauper:09, Liu:18}. \newcite{Sauper:09} induce domain-specific templates from Wikipedia and fill these templates with content from the Internet. More recently \newcite{Liu:18} explore a diverse set of domains for summarization and are the first to attempt abstractive summarization of the first section of Wikipedia articles, by combining extractive and abstractive summarization methods.
\vspace{-.1cm}
\section{Dataset Collection}
\vspace{-.1cm}
\subsection{An Overview of TutorialBank}
 As opposed to other collections like the ACL Anthology \cite{Radev:08, Radev:09,Radev:13, Radev:16}, which contain solely academic papers, our corpus focuses mainly on resources other than academic papers. The main goal in our decision process of what to include in our corpus has been the quality-control of resources which can be used for an educational purpose. Initially, the resources collected were conference tutorials as well as surveys, books and longer papers on broader topics, as these genres contain an inherent amount of quality-control. Later on, other online resources were added to the corpus, as explained below. Student annotators, described later on, as well as the professor examined resources which they encountered in their studies. The resources were added to the corpus if deemed of good quality. Important to note is that not all resources which were found on the Internet were added to TutorialBank; one could scrape the web according to search terms, but quality control of the results would be largely missing. The quality of a resource is a somewhat subjective measure, but we aimed to find resources which would serve a pedagogical function to either students or researchers, with a professor of NLP making the final decision. This collection of resources and meta-data annotation has been done over multiple years, while this year we created the search engine and added additional annotations mentioned below.

\begin{table}[!t]
\small 
    \centering
\begin{tabular}{ |p{7cm}|}
 \hline
1 - Introduction and Linguistics\\
 \hline
 2 - Language Modeling, Syntax and Parsing \\
 \hline
 3 - Semantics and Logic \\
 \hline
  4 - Pragmatics, Discourse, Dialogue and Applications \\
 \hline
  5 - Classification and Clustering\\
 \hline
  6 - Information Retrieval and Topic Modeling \\
 \hline
   7 - Neural Networks and Deep Learning \\
 \hline
   8 - Artificial Intelligence \\
     \hline
   9 - Other Topics \\
 \hline
\end{tabular}
    \caption{Top-level Taxonomy Topics}
    \label{tab:tax}
\end{table}

\begin{table}
\small
    \centering
\begin{tabular}{ |M{5cm}||M{1.5cm}| }
 \hline
Topic Category & Count \\ 
\hline
Introduction to Neural Networks and Deep Learning & 635\\
 \hline
Tools for Deep Learning & 475\\
 \hline
 Miscellaneous Deep Learning & 287\\
 \hline
  Machine Learning & 225\\
   \hline
  Word Embeddings & 139\\

 \hline
  Recurrent Neural Networks & 134\\
 \hline
   Python Basics & 133\\
 \hline
   Reinforcement learning & 132\\
 \hline
    Convolutional Neural Networks & 129\\
 \hline
    Introduction to AI & 89\\
 \hline
\end{tabular}
    \caption{Corpus count by taxonomy topic for the most frequent topics (excluding topic ``Other").}
    \vspace{-.3cm}
    \label{tab:topics}
\end{table}

\subsubsection{TutorialBank Taxonomy}
In order to facilitate the sharing of resources about NLP, we developed a taxonomy of 305 topics of varying granularity. The top levels of our taxonomy tree are shown in Table \ref{tab:tax}. The backbone of our Taxonomy corresponds to the syllabus of a university-level NLP course and was expanded to include related topics from other courses in ML, IR and AI. As a result, there is a bias in the corpus towards NLP resources and resources from other fields in so far as they are relevant to NLP. However, this bias is planned, as our focus remains teaching NLP. The resource count for the most frequent taxonomy topics is shown in Table \ref{tab:topics}.

\subsection{Data Preprocessing}
For each resource in the corpus, we downloaded the corresponding PDF, PowerPoint presentations and other source formats and used PDFBox to perform OCR in translating the files to textual format. For HTML pages we downloaded both the raw HTML with all images as well as a formatted text version of the pages. For copyright purposes we release only the meta data such as urls and annotations and provide scripts for reproducing the dataset.

\section{Dataset Annotation}
Annotations were performed by a group of 3 PhD students in NLP, and 6 undergraduate Computer Science students who have taken at least one course in AI or NLP. 
\subsection{Pedagogical Function}
When collecting resources from the Internet, each item was labeled according to the medium in which it was found, analogous to the pedagogical function of \cite{Sheng:17}. We will use this term throughout the paper to describe this categorization. The categories along with their counts are shown in Table \ref{tab:pedagogicalcounts}: 

\begin{description}
\itemsep.03em 
  \item[$\bullet$ Corpus:] A corpus provides access to and a description of a scientific dataset.  
  \item[$\bullet$ Lecture:] A lecture consists of slides/notes from a university lecture.
    \item[$\bullet$ Library:] A library consists of github pages and other codebases which aid in the implementation of algorithms. 
  \item[$\bullet$ NACLO:] NACLO problems refer to linguistics puzzles from the North American Computational Linguistics Olympiad. 
  \item[$\bullet$ Paper:] A paper is a short/long conference paper taken from sites such as https://arxiv.org/ and which is not included in the ACL Anthology.  
  \item[$\bullet$ Link set:] A link set provides a collection of helpful links in one location. 
  \item[$\bullet$ Survey:] A survey is a long paper or book which describes a broader subject. 
  \item[$\bullet$ Tutorial:] A tutorial is a slide deck from a conference tutorial or an HTML page that describes a contained topic. 
\end{description}

\begin{table}
\small
    \centering
\begin{tabular}{ |M{3cm}||M{2cm}| }
 \hline
Resource Category & Count \\ 
\hline 
corpus & 131\\
 \hline
 lecture & 126\\
 \hline
  library & 1014\\
   \hline
  link set & 1186\\
 \hline
   naclo & 154\\
 \hline
  paper & 1176\\
 \hline
   survey & 390\\
 \hline
    tutorial & 2079\\
 \hline
\end{tabular}
    \caption{Corpus count by pedagogical feature.}
    \label{tab:pedagogicalcounts}
\end{table}

\subsection{Topic to Resource Collection}
We first identified by hand 200 potential topics for survey generation in the fields of NLP, ML, AI and IR. Topics were added according to the following criteria:
\begin{enumerate}
  \item  It is conceivable that someone would write a Wikipedia page on this topic (an actual page may or may not exist).
  \item The topic is not overly general (e.g., ``Natural Language Processing") or too obscure or narrow.
  \item  In order to write a survey on the topic, one would need to include information from a number of sources.
\end{enumerate}

\begin{table}
\small
    \centering
\begin{tabular}{ |M{5cm}| }
\hline 
Capsule Networks \\
 \hline
Domain Adaptation \\
 \hline
 Document Representation\\
 \hline
 Matrix factorization\\
 \hline
 Natural language generation \\
 \hline
  Q Learning \\
 \hline
  Recursive Neural Networks\\
 \hline
    Shift-Reduce Parsing \\
    \hline
    Speech Recognition \\
 \hline
 Word2Vec \\
 \hline
\end{tabular}
    \caption{Random sample of the list of 200 topics used for prerequisite chains, readling lists and survey extraction.}
    \label{tab:200topics}
\end{table}

While some of the topics come from our taxonomy, many of the taxonomy topics have a different granularity than we desired, which motivated our topic collection. Topics were added to the list along with their corresponding Wikipedia pages, if they exist. A sample of the topics selected is shown in \ref{tab:200topics}. Once the list of topics was compiled, annotators were assigned topics and asked to search that topic in the TutorialBank search engine and find relevant resources. 
In order to impose some uniformity on the dataset, we chose to only include resources which consisted of PowerPoint slides as well as HTML pages labeled as tutorials. 
We divided the topics among the annotators and asked them to choose five resources per topic using our search engine. The resource need not solely focus on the given topic; the resource may be on a more general topic and include a section on the given topic. As in general searching for resources, often resources include related information, so we believe this setting is fitting. For some topics the annotators chose fewer than five resources (partially due to the constraint we impose on the form of the resources). We noted topics for which no resources were found, and rather than replace the topics to reflect TutorialBank coverage, we leave these topics in and plan to add additional resources in a future release. 

\subsection{Prerequisite Chains}
Even with a collection of resources and a list of topics, a student may not know where to begin studying a topic of interest. For example, in order to understand sentiment analysis the student should be familiar with Bayes' Theorem, the basics of ML as well as other topics. For this purpose, the annotators annotated which topics are prerequisites of others for the given topics from their reading lists. We expanded our list of potential prerequisites to include  eight additional topics which were too broad for survey generation (e.g., Linear Algebra) but which are important prerequisites to capture. Following the method of \cite{Gordon:16}, we define labeling a topic Y as a prerequisite of X according to the following question:
\begin{description}
\itemsep.03em 
  \item[$\bullet$] Would understanding Topic Y help you to understand Topic X?
\end{description}
 As in \cite{Gordon:16}, the annotators can answer this question as ``no", ``somewhat" or ``yes." 
\subsection{Reading Lists}
When annotators were collecting relevant resources for a particular topic, we asked them to order the resources they found in terms of the usefulness of the resource for learning that particular topic. We also include the Wikipedia pages corresponding to the topics, when available, as an additional source of information. We do not perform additional annotation of the order of the resources or experiment in automatically reproducing these ordered lists but rather offer this annotation as a pedagogical tool for students and educators. We plan the expansion of these lists and analysis in future experiments. 

\subsection{Survey Extraction}
We frame the task of creating surveys of scientific topics as a document retrieval task. A student searching for resources in order to learn about a topic such as Recurrent Neural Networks (RNN's) may encounter resources 1) which solely cover RNN's as well as 2) resources which cover RNN's within the context of a larger topic (e.g., Deep Learning). Within the first type, not every piece of content (a single PowerPoint slide or section in a blog post) contributes equally well to an understanding of RNN's; the content may focus on background information or may not clearly explain the topic. Within the second type, larger tutorials may contain valuable information on the topic, but may also contain much information not immediately relevant to the query. Given a query topic and a set of parsed documents we want to retrieve the parts most relevant to the topic. 
\par
In order to prepare the dataset for extracting surveys of topics, we first divide resources into units of content which we call ``cards". PowerPoint slides inherently contain a division in the form of each individual slide, so we divide PowerPoint presentations into individual slides/cards. For HTML pages, the division is less clear. However, we convert the HTML pages to a markdown file and then automatically split the markdown file using header markers. We believe this is a reasonable heuristic as tutorials and similar content tend to be broken up into sections signalled by headers.
\par
For each of the resources which the annotators gathered for the reading lists on a given topic, that same annotator was presented with each card from that resource and asked to rate the usefulness of the card. The annotator could rate the card from 0-2, with 0 meaning the card is not useful for learning the specified topic, 1 meaning the card is somewhat useful and 2 meaning the card is useful. We chose a 3-point scale as initial trials showed a 5-point scale to be too subjective. The annotators also had the option in our annotation interface to drop cards which were parsed incorrectly or were repeated one after the other as well as skip cards and return to score a card.

\subsection{Illustrations}
Whether needed for understanding a subject more deeply or for preparing a blog post on a subject, images play an important role in presenting concepts more concretely. Simply extracting the text from HTML pages leaves behind this valuable information, and OCR software often fails to parse complex graphs and images in a non-destructive fashion. To alleviate this problem and promote the sharing of images, we extracted all images from our collected HTML pages. Since many images were simply HTML icons and other extraneous images, we manually checked the images and selected those which are of value to the NLP student. We collected a total of 2,000 images and matched them with the taxonomy topic name of the resource it came from as well as the url of the resource. While we cannot outdo the countless images from Google search, we believe illustrations can be an additional feature of our search engine, and we describe an interface for this collection below. 

\section{Additional Features and Analysis}
\subsection{Search Engine}
In order to present our corpus in a user-friendly manner, we created a search engine using Apache Lucene\footnote{http://lucene.apache.org/}. We allow the user to query key words to search our resource corpus, and the results can then be sorted based on relevance, year, topic, medium, and other meta data. In addition to searching by term, users can browse the resources by topic according to our taxonomy. For each child topic from the top-level taxonomy downward, we display resources according to their pedagogical functions. In addition to searching for general resources, we also provide search functionality for a corpus of papers, where the user can search by keyword as well as by author and venue.
\par
While the search engine described above provides access to our base corpus and meta data, we also provide a command-line interface tool with our release  so that students and researchers can easily use our annotations for prerequisite topics, illustrations and survey generation for educational purposes. The tool allows the user to input a topic from the taxonomy and retrieve all images related to that topic according to our meta data. Additionally, the user can input a topic from our list of 200 topics, and our tool outputs the prerequisites of that topic according to our annotation as well as the cards labelled as relevant for that topic. 
\subsection{Resource Recommendation from Title and Abstract Pairs}
In addition to needing to search for a general term, often a researcher begins with an idea for a project which is already focused on a nuanced sub-task. An employee at an engineering company may be starting a project on image captioning. Ideas about the potential direction of this project may be clear, but what resources may be helpful or what papers have already been published on the subject may not be immediately obvious. To this end we propose the task of recommending resources from title and abstract pairs. The employee will input the title and abstract of the project and obtain a list of resources which can help complete the project. This task is analogous to reproducing the reference section of a paper, however, with a focus on tutorials and other resources rather than solely on papers. As an addition to our search engine, we allow a user to input a title and an abstract of variable length. We then propose taxonomy topics based on string matches with the query as well as a list of resources and papers and their scores as determined by the search engine. We later explore two baseline models for recommending resources based on document and topic modeling.

\begin{figure*}[!htb]
  \includegraphics[width=6in]{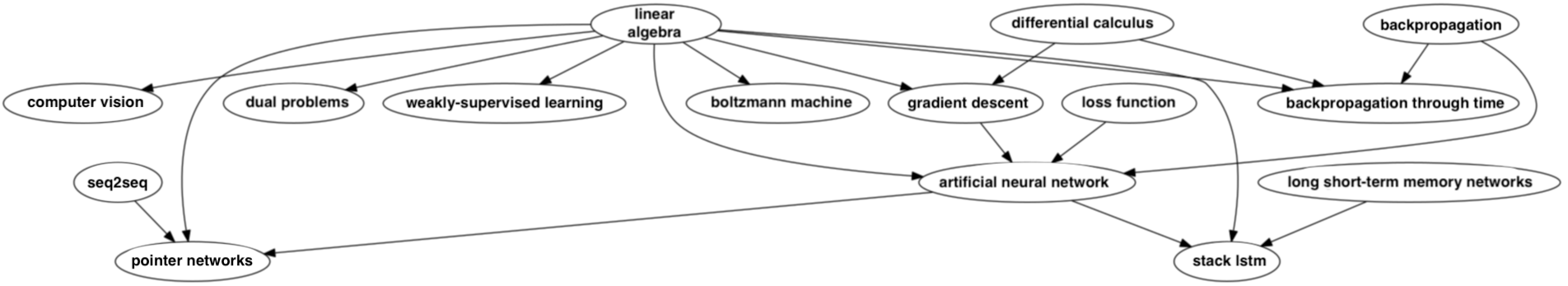}
  \caption{Subset of prerequisite annotations taken from inter-annotator agreement round.}
  \label{fig:prereqs}
\end{figure*}

\subsection{Dataset and Annotation Statistics}
We created reading lists for 182 of the 200 topics we identify in Section 4.2. Resources were not found for 18 topics due to the granularity of the topic (e.g., Radial Basis Function Networks) as well as our intended restriction of the chosen resources to PowerPoint presentations and HTML pages. The average number of resources per reading list for the 182 topics is 3.94. As an extension to the reading lists we collected Wikipedia pages for 184 of the topics and present these urls as part of the dataset. 
\par
We annotated prerequisite relations for the 200 topics described above. We present a subset of our annotations in Figure \ref{fig:prereqs}, which shows the network of topic relations  (nodes without incoming edges were not annotated for their prerequisites as part of this shown inter-annotation round). Our network consists of 794 unidirectional edges and 33 bidirectional edges. The presence of bidirectional edges stems from our definition of a prerequisite, which does not preclude bidirectionality (one topic can help explain another and vice-versa) as well as the similarity of the topics. The set of bidirectional edges consists of topic pairs (BLEU - ROUGE; Word Embedding - Distributional Semantics; Backpropagation - Gradient descent) which could be collapsed into one topic to create a directed acyclic graph in the future. 
\par
For survey extraction, we automatically split 313 resources into content cards which we annotated for usefulness in survey extraction. These resources are a subset of the reading lists limited in number due to constraints in downloading urls and parsing to our annotation interface. The total number of cards which were not marked as repeats/mis-parsed totals 17,088, with 54.59 per resource. 6,099 cards were labeled as somewhat relevant or relevant for the target topic. The resources marked as non-relevant may be poorly presented or may not pertain fully to the topic of that survey. These numbers confirm the appropriateness of this survey corpus as a non-trivial information retrieval task.
\par
To better understand the difficulty of our annotation tasks, we performed inter-annotator agreement experiments for each of our annotations. We randomly sampled twenty-five resources and had annotators label for pedagogical function. Additionally, we sampled twenty-five  topics for prerequisite annotations and five  topics with reading list lengths of five for survey annotation.  We used Fleiss's Kappa \cite{Fleis:2004}, a variant of Cohen's Kappa \cite{Cohen:60} designed to measure annotator agreement for more than two annotators. The results are shown in Table \ref{tab:inter_annotator}. Using the scale as defined in \newcite{Landis:1977}, pedagogical function annotation exhibits \textit{substantial agreement} while prerequisite annotation and survey extraction annotation show \textit{fair agreement}. The Kappa score for pedagogical function is comparable to that of \newcite{Sheng:17} (0.68) while the prerequisite annotation is slightly lower than the agreement metric used in \newcite{Gordon:16} (0.36) although they measure agreement through Pearson correlation. We believe that the sparsity of the labels plays a role in these scores. 

\begin{table}
\small
    \centering
\begin{tabular}{ |M{4cm}||M{2.5cm}| }
 \hline
Annotation & Kappa \\
\hline
Pedagogical Function & 0.69 \\
\hline
Prerequisites & 0.30\\
\hline
Survey Extraction & 0.33 \\
\hline
\end{tabular}
    \caption{Inter-annotator agreement.}
    \label{tab:inter_annotator}
      \vspace{-.3cm}
\end{table}



\begin{figure*}[!htb]
    \centerline{\resizebox{1\textwidth}{!}{\includegraphics{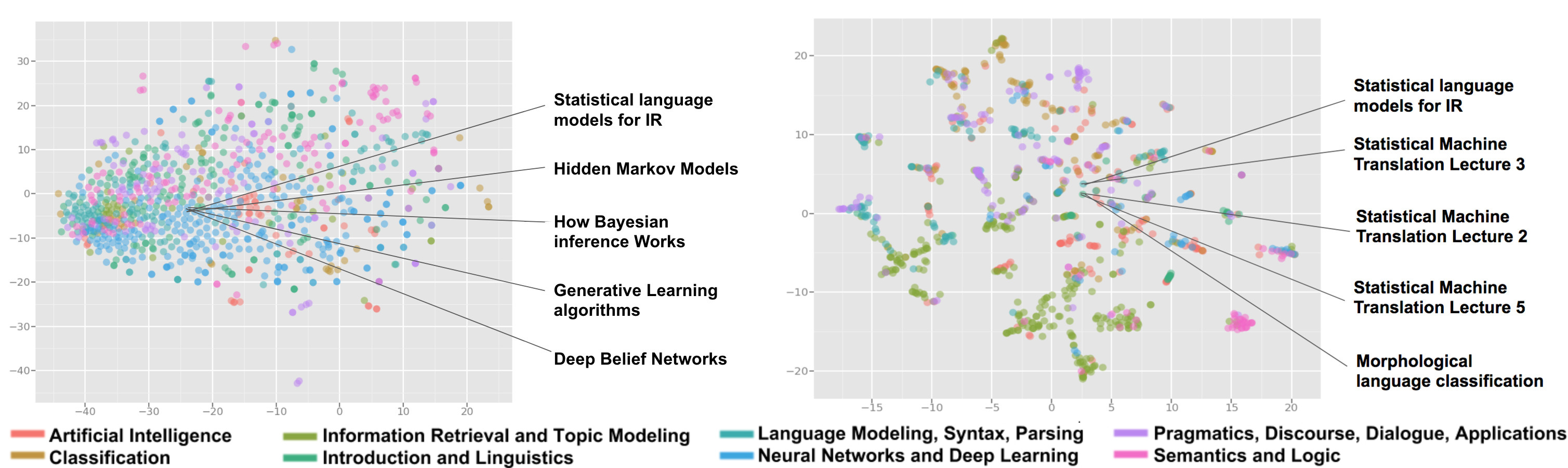}} }

  \caption{Plot showing a query document with title ``Statistical language models for IR" and its neighbour document clusters as obtained through tSNE dimension reduction for Doc2Vec (left) and LDA topic modeling (right). Nearest neighbor documents titles are shown to the right of each plot.}
  \vspace{-.3cm}
  \label{fig:clusters}
\end{figure*}

\subsection{Comparison to Similar Datasets}
Our corpus distinguishes itself in its magnitude, manual collection and focus on annotation for educational purposes in addition to research tasks. 
We use similar categories for classifying pedagogical function as \newcite{Sheng:17}, but our corpus is hand-picked and over four-times larger, while exhibiting similar annotation agreement. 
\par
\newcite{Gordon:16} present a corpus for prerequisite relations among topics, but this corpus differs in coverage. They used LDA topic modeling to generate a list of 300 topics, while we manually create a list of 200 topics based on criteria described above. Although their topics are generated from the ACL Anthology and related to NLP, we find less than a 40\% overlap in topics. Additionally, they only annotate a subset of the topics for prerequisite annotations while we focus on broad coverage, annotating two orders of magnitude larger in terms of prerequisite edges while exhibiting fair inter-annotator agreement.  
\par
Previous work and datasets on generating surveys for scientific topics have focused on scientific articles \cite{Jha:13, Jha:15, Jaidka:16} and Wikipedia pages \cite{Sauper:09, Liu:18} as a summarization task. We, on the other hand, view this problem as an information retrieval task and focus on extracting content from manually-collected PowerPoint slides and online tutorials. \newcite{Sauper:09} differ in their domain coverage, and while the surveys of \newcite{Jha:13, Jha:15} focus on NLP, we collect resources for an order of magnitude larger set of topics. Finally, our focus here in creating surveys, as well as the other annotations, is first and foremost to create a useful tool for students and researchers. Websites such as the ACL Anthology\footnote{http://aclweb.org/anthology/} and arXiv\footnote{https://arxiv.org/} provide an abundance of resources, but do not focus on the pedagogical aspect of their content. Meanwhile, websites such as Wikipedia which aim to create a survey of a topic may not reflect the latest trends in rapidly changing fields.

\section{Topic Modeling and Resource Recommendation}
As an example usage of our corpus, we experimented with topic modeling and its extension to resource recommendation. We restricted our corpus for this study to non-HTML files to examine the single domain of PDF's and PowerPoint presentations. This set consists of about 1,480 files with a vocabulary size  191,446 and a token count of 9,134,452. For each file, the tokens were processed, stop tokens were stripped, and then each token was stemmed. Words with counts less than five across the entire corpus were dropped. We experimented with two models: LDA, a generative probabilistic model mentioned earlier, and Doc2Vec \cite{Le:14}, an extension of Word2Vec \cite{Mikolov:13} which creates representations of arbitrarily-sized documents. Figure \ref{fig:clusters} shows the document representations obtained with Doc2Vec as well as the topic clusters created with LDA. The grouping of related resources around a point demonstrates the clustering abilities of these models. We applied LDA in an unsupervised way, using 60 topics over 300 iterations as obtained through experimentation, and then colored each document dot with its category to observe the distribution. Our Doc2Vec model used hidden dimension 300, a window size of 10 and a constant learning rate of 0.025. Then, the model was trained for 10 epochs.  
\par
\begin{figure}
    \centering
  \begin{tikzpicture}[scale=0.50]
    \begin{axis}[
        width  = \textwidth,
        height = 8cm,
        major x tick style = transparent,
        ybar=5*\pgflinewidth,
        bar width=14pt,
        ymajorgrids = true,
        ylabel = \textbf{Percent of relevant recommendations},
        xlabel = \textbf{Test Document ID},
        symbolic x coords={1,2,3,4,5,6,7,8,9,10},
        xtick = data,
        scaled y ticks = false,
        ymin=0,
        ymax=1,
        legend columns=2,
        legend cell align=left,
        legend style={
                at={(1,.9)},
                anchor=south east,
                column sep=1ex
        }
    ]
        \addplot[style={blue,fill=blue,mark=none}]
          coordinates {(1, .21) (2, .57) (3, .32) (4, .39) (5, .41) (6,.27) (7,.20) (8,.64) (9,.18)(10,.18)};

        \addplot[style={red,fill=red,mark=none}]
             coordinates {(1, 0.46 ) (2,  0.42 ) (3, 0.39  ) (4,  0.42  ) (5, 0.68  ) (6,  0.83) (7,0.18) (8, 0.28 )(9, 0.47  )(10, 0.39 )};
               \legend{\textbf{Doc2Vec}, \textbf{LDA}}
    \end{axis}
\end{tikzpicture}  
    \caption{Relevance accuracies of the Doc2Vec and LDA resource recommendation models.}
    \vspace{-.4cm}
    \label{fig:recs}
\end{figure}
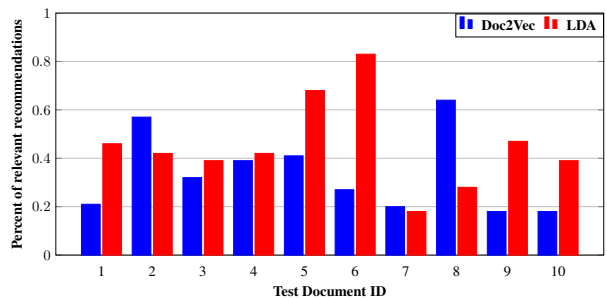

We tested these models for the task of resource recommendation from title+abstract pairs.  We collected 10 random papers from ACL 2017. For LDA, the document was classified to a topic, and then the top resources from that topic were chosen, while Doc2Vec computed the similarity between the query document and the training set and chose the most similar documents. We concatenated the title and abstract as input and had our models predict the top 20 documents. We then had five annotators rate the recommendations for helpfulness as 0 (not helpful) or 1 (helpful). Recommended resources were rated according to the criterion of whether reading this resource would be useful in doing a project as described in the title and abstract. The results are found in Figure \ref{fig:recs}. Averaging the performance over each test case, the LDA model performed better than Doc2Vec (0.45 to 0.34), although both leave large room for improvements. LDA recommended resources notably better for cases 5 and 6, which correspond to papers with very well defined topics areas (Question Answering and Machine Translation) while Doc2Vec was able to find similar documents for cases 2 and 8 which are a mixture of topics, yet are well-represented in our corpus (Reinforcement Learning with dialog agents and emotion (sentiment) detection with classification). The low performance for both models also corresponds to differences in corpus coverage, and we plan to explore this bias in the future. We believe that this variant of reading list generation as well as the relationship between titles and abstracts is an unexplored and exciting area for future research. 

\section{Conclusion and Future Work}
In this paper we introduce the TutorialBank Corpus, a collection of over 6,300 hand-collected resources on NLP and related fields. Our corpus is notably larger than similar datasets which deal with pedagogical resources and topic dependencies and unique in use as an educational tool. To this point, we believe that this dataset, with its multiple layers of annotation and usable interface, will be an invaluable tool to the students, educators and researchers of NLP. Additionally, the corpus promotes research on tasks not limited to pedagogical function classification, topic modeling and prerequisite relation labelling. Finally, we formulate the problem of recommending resources for a given title and abstract pair as a new way to approach reading list generation and propose two baseline models. For future work we plan to continue the collection and annotation of resources and to separately explore each of the above research tasks.  


\section*{Acknowledgments}
We would like to thank all those who worked on the development of the search engine and website as well as those whose discussion and annotated greatly helped this work, especially Jungo Kasai,  Alexander Strzalkowski, Michihiro Yasunaga and Rui Zhang. 

\bibliography{acl2018}

\begin{thebibliography}{25}
\expandafter\ifx\csname natexlab\endcsname\relax\def\natexlab#1{#1}\fi

\bibitem[{Bird et~al.(2008)Bird, Dale, Dorr, Gibson, Joseph, Kan, Lee, Powley,
  Radev, and Tan}]{Radev:08}
Steven Bird, Robert Dale, Bonnie~J. Dorr, Bryan~R. Gibson, Mark~Thomas Joseph,
  Min{-}Yen Kan, Dongwon Lee, Brett Powley, Dragomir~R. Radev, and Yee~Fan Tan.
  2008.
\newblock The {ACL} {A}nthology {R}eference {C}orpus: {A} {R}eference {D}ataset
  for {B}ibliographic {R}esearch in {C}omputational {L}inguistics.
\newblock In \emph{{LREC}}. European Language Resources Association.

\bibitem[{Blei et~al.(2003)Blei, Ng, and Jordan}]{Blei:03}
David~M. Blei, Andrew~Y. Ng, and Michael~I. Jordan. 2003.
\newblock Latent {D}irichlet {A}llocation.
\newblock \emph{Journal of Machine Learning Research}, 3:993--1022.

\bibitem[{Cohen(1960)}]{Cohen:60}
Jacob Cohen. 1960.
\newblock A {C}oefficient of {A}greement for {N}ominal {S}cales.
\newblock \emph{Educational and psychological measurement}, 20(1):37--46.

\bibitem[{Fleiss et~al.(2004)Fleiss, Levin, and Paik}]{Fleis:2004}
Joseph~L. Fleiss, Bruce Levin, and Myunghee~Cho Paik. 2004.
\newblock \href {https://doi.org/10.1002/0471445428.ch18} {\emph{The
  {M}easurement of {I}nterrater {A}greement}}. John Wiley \& Sons, Inc.

\bibitem[{Gordon et~al.(2017)Gordon, Aguilar, Sheng, and Burns}]{Gordon:17}
Jonathan Gordon, Stephen Aguilar, Emily Sheng, and Gully Burns. 2017.
\newblock Structured {G}eneration of {T}echnical {R}eading {L}ists.
\newblock In \emph{BEA@EMNLP}, pages 261--270. Association for Computational
  Linguistics.

\bibitem[{Gordon et~al.(2016)Gordon, Zhu, Galstyan, Natarajan, and
  Burns}]{Gordon:16}
Jonathan Gordon, Linhong Zhu, Aram Galstyan, Prem Natarajan, and Gully Burns.
  2016.
\newblock Modeling {C}oncept {D}ependencies in a {S}cientific {C}orpus.
\newblock In \emph{Proceedings of the 54th Annual Meeting of the Association
  for Computational Linguistics, {ACL} 2016, August 7-12, 2016, Berlin,
  Germany, Volume 1: Long Papers}.

\bibitem[{Jaidka et~al.(2016)Jaidka, Chandrasekaran, Rustagi, and
  Kan}]{Jaidka:16}
Kokil Jaidka, Muthu~Kumar Chandrasekaran, Sajal Rustagi, and Min{-}Yen Kan.
  2016.
\newblock Overview of the {C}l-{S}ci{S}umm 2016 {S}hared {T}ask.
\newblock In \emph{BIRNDL@JCDL}, volume 1610 of \emph{{CEUR} Workshop
  Proceedings}, pages 93--102. CEUR-WS.org.

\bibitem[{Jardine(2014)}]{Jardine:14}
James~Gregory Jardine. 2014.
\newblock \emph{Automatically {G}enerating {R}eading {L}ists}.
\newblock Ph.D. thesis, University of Cambridge, {UK}.

\bibitem[{Jha et~al.(2013)Jha, Abu{-}Jbara, and Radev}]{Jha:13}
Rahul Jha, Amjad Abu{-}Jbara, and Dragomir~R. Radev. 2013.
\newblock A {S}ystem for {S}ummarizing {S}cientific {T}opics {S}tarting from
  {K}eywords.
\newblock In \emph{Proceedings of the 51st Annual Meeting of the Association
  for Computational Linguistics, {ACL} 2013, 4-9 August 2013, Sofia, Bulgaria,
  Volume 2: Short Papers}, pages 572--577.

\bibitem[{Jha et~al.(2015)Jha, Coke, and Radev}]{Jha:15}
Rahul Jha, Reed Coke, and Dragomir~R. Radev. 2015.
\newblock Surveyor: {A} {S}ystem for {G}enerating {C}oherent {S}urvey
  {A}rticles for {S}cientific {T}opics.
\newblock In \emph{Proceedings of the Twenty-Ninth {AAAI} Conference on
  Artificial Intelligence, January 25-30, 2015, Austin, Texas, {USA.}}, pages
  2167--2173.

\bibitem[{Landis and Koch(1977)}]{Landis:1977}
J~Richard Landis and Gary~G Koch. 1977.
\newblock The {M}easurement of {O}bserver {A}greement for {C}ategorical {D}ata.
\newblock \emph{Biometrics}, pages 159--174.

\bibitem[{Le and Mikolov(2014)}]{Le:14}
Quoc~V. Le and Tomas Mikolov. 2014.
\newblock Distributed {R}epresentations of {S}entences and {D}ocuments.
\newblock \emph{CoRR}, abs/1405.4053.

\bibitem[{Liang et~al.(2017)Liang, Ye, Wu, Pursel, and Giles}]{Liang:17}
Chen Liang, Jianbo Ye, Zhaohui Wu, Bart Pursel, and C.~Lee Giles. 2017.
\newblock Recovering {C}oncept {P}rerequisite {R}elations from {U}niversity
  {C}ourse {D}ependencies.
\newblock In \emph{Proceedings of the Thirty-First {AAAI} Conference on
  Artificial Intelligence, February 4-9, 2017, San Francisco, California,
  {USA.}}, pages 4786--4791.

\bibitem[{Liu et~al.(2016)Liu, Ma, Yang, and Carbonell}]{Liu:16}
Hanxiao Liu, Wanli Ma, Yiming Yang, and Jaime~G. Carbonell. 2016.
\newblock Learning {C}oncept {G}raphs from {O}nline {E}ducational {D}ata.
\newblock \emph{J. Artif. Intell. Res.}, 55:1059--1090.

\bibitem[{Liu et~al.(2018)Liu, Saleh, Pot, Goodrich, Sepassi, Kaiser, and
  Shazeer}]{Liu:18}
Peter~J. Liu, Mohammad Saleh, Etienne Pot, Ben Goodrich, Ryan Sepassi, Lukasz
  Kaiser, and Noam Shazeer. 2018.
\newblock \href {https://openreview.net/forum?id=Hyg0vbWC-} {Generating
  {W}ikipedia by {S}ummarizing {L}ong {S}equences}.
\newblock \emph{International Conference on Learning Representations}.

\bibitem[{Mikolov et~al.(2013)Mikolov, Chen, Corrado, and Dean}]{Mikolov:13}
Tomas Mikolov, Kai Chen, Greg Corrado, and Jeffrey Dean. 2013.
\newblock Efficient {E}stimation of {W}ord {R}epresentations in {V}ector
  {S}pace.
\newblock \emph{CoRR}, abs/1301.3781.

\bibitem[{Pan et~al.(2017{\natexlab{a}})Pan, Li, Li, and Tang}]{Pan:17}
Liangming Pan, Chengjiang Li, Juanzi Li, and Jie Tang. 2017{\natexlab{a}}.
\newblock \href {https://doi.org/10.18653/v1/P17-1133} {Prerequisite {R}elation
  {L}earning for {C}oncepts in {MOOC}s}.
\newblock In \emph{Proceedings of the 55th Annual Meeting of the Association
  for Computational Linguistics, {ACL} 2017, Vancouver, Canada, July 30 -
  August 4, Volume 1: Long Papers}, pages 1447--1456.

\bibitem[{Pan et~al.(2017{\natexlab{b}})Pan, Wang, Li, Li, and Tang}]{Pan:17b}
Liangming Pan, Xiaochen Wang, Chengjiang Li, Juanzi Li, and Jie Tang.
  2017{\natexlab{b}}.
\newblock Course {C}oncept {E}xtraction in {MOOC}s via {E}mbedding-{B}ased
  {G}raph {P}ropagation.
\newblock In \emph{{IJCNLP(1)}}, pages 875--884. Asian Federation of Natural
  Language Processing.

\bibitem[{Radev et~al.(2016)Radev, Joseph, Gibson, and
  Muthukrishnan}]{Radev:16}
Dragomir~R. Radev, Mark~Thomas Joseph, Bryan~R. Gibson, and Pradeep
  Muthukrishnan. 2016.
\newblock A {B}ibliometric and {N}etwork {A}nalysis of the {F}ield of
  {C}omputational {L}inguistics.
\newblock \emph{{JASIST}}, 67(3):683--706.

\bibitem[{Radev et~al.(2009)Radev, Muthukrishnan, and Qazvinian}]{Radev:09}
Dragomir~R. Radev, Pradeep Muthukrishnan, and Vahed Qazvinian. 2009.
\newblock The {ACL} {A}nthology {N}etwork {C}orpus.
\newblock In \emph{Proceedings, ACL Workshop on Natural Language Processing and
  Information Retrieval for Digital Libraries}, Singapore.

\bibitem[{Radev et~al.(2013)Radev, Muthukrishnan, Qazvinian, and
  Abu{-}Jbara}]{Radev:13}
Dragomir~R. Radev, Pradeep Muthukrishnan, Vahed Qazvinian, and Amjad
  Abu{-}Jbara. 2013.
\newblock \href {https://doi.org/10.1007/s10579-012-9211-2} {The {ACL}
  {A}nthology {N}etwork {C}orpus}.
\newblock \emph{Language Resources and Evaluation}, 47(4):919--944.

\bibitem[{Sauper and Barzilay(2009)}]{Sauper:09}
Christina Sauper and Regina Barzilay. 2009.
\newblock Automatically {G}enerating {W}ikipedia {A}rticles: {A}
  {S}tructure-{A}ware {A}pproach.
\newblock In \emph{{ACL} 2009, Proceedings of the 47th Annual Meeting of the
  Association for Computational Linguistics and the 4th International Joint
  Conference on Natural Language Processing of the AFNLP, 2-7 August 2009,
  Singapore}, pages 208--216.

\bibitem[{Sheng et~al.(2017)Sheng, Natarajan, Gordon, and Burns}]{Sheng:17}
Emily Sheng, Prem Natarajan, Jonathan Gordon, and Gully Burns. 2017.
\newblock An {I}nvestigation into the {P}edagogical {F}eatures of {D}ocuments.
\newblock In \emph{BEA@EMNLP}, pages 109--120. Association for Computational
  Linguistics.

\bibitem[{Tang and McCalla(2004)}]{Tang:04}
Tiffany~Ya Tang and Gordon~I. McCalla. 2004.
\newblock On the {P}edagogically {G}uided {P}aper {R}ecommendation for an
  {E}volving {W}eb-{B}ased {L}earning {S}ystem.
\newblock In \emph{{FLAIRS} Conference}, pages 86--92. {AAAI} Press.

\bibitem[{Tang and McCalla(2009)}]{Tang:09}
Tiffany~Ya Tang and Gordon~I. McCalla. 2009.
\newblock The {P}edagogical {V}alue of {P}apers: a {C}ollaborative-{F}iltering
  based {P}aper recommender.
\newblock \emph{J. Digit. Inf.}, 10(2).

\end{thebibliography}
\bibliographystyle{acl2018}



 




\end{document}